\documentclass[a4paper, 10pt, conference]{IEEEtran}
\IEEEoverridecommandlockouts
\usepackage[a4paper, left=0.625in, right=0.625in, bottom=1in, top=0.75in]{geometry}
\usepackage{cite}
\usepackage{amsmath,amssymb,amsfonts}
\usepackage{algorithmicx}
\usepackage{graphicx}
\usepackage{textcomp}
\usepackage{todonotes}
\usepackage{duckuments}
\usepackage{lipsum}
\usepackage{tabularx}
\usepackage{booktabs}
\usepackage{multirow}
\usepackage{slantsc}
\usepackage{xltabular}
\usepackage{float}
\usepackage{algorithm}
\usepackage{algpseudocode}
\usepackage{tabularray}
\usepackage{multicol}
\usepackage{stfloats}
\usetikzlibrary{shapes}
\usetikzlibrary{shapes.geometric}
\usepackage{svg}
\usepackage[section]{placeins}
\usepackage[caption=false]{subfig}
\usepackage[nolist]{acronym}
\usepackage{hyperref}
\usepackage{xcolor}
\usepackage[bottom]{footmisc}

\newcommand{\tabitem}{~~\llap{\textbullet}~~}

\include{content/definitions}

\usetikzlibrary{arrows,chains,positioning,fit}
\newcommand{\krug}[1]{\tikz[baseline=(char.base)]{\node[shape=circle,draw,minimum size=4mm, inner sep=1pt, semithick](char){#1}}}
\newcommand{\kvadrat}[1]{\tikz[baseline=(char.base)]{\node[shape=rectangle,draw,minimum size=4mm, inner sep=1pt, semithick](char){#1}}}

\makeatletter
\newlength{\trianglerightwidth}
\algnewcommand{\LineComment}[1]{\Statex \hskip\ALG@thistlm $\triangleright$ #1}
\algnewcommand{\LineCommentCont}[1]{\Statex \hskip\ALG@thistlm%
	\parbox[t]{\dimexpr\linewidth-\ALG@thistlm}{\hangindent=\trianglerightwidth \hangafter=1 \strut$\triangleright$ #1\strut}}
\makeatother

\begin{document}
	\title{Dynamic Collaborative Path Planning for Remote Assistance of Highly-Automated Vehicles}

\author{\IEEEauthorblockN{Domagoj Majstorovi\'c}
	\IEEEauthorblockA{\textit{Institute of Automotive Technology} \\
		\textit{Technical University of Munich}\\
		Garching bei M{\"u}nchen, Germany \\
		domagoj.majstorovic@tum.de}
	\and
	\IEEEauthorblockN{Frank Diermeyer}
	\IEEEauthorblockA{\textit{Institute of Automotive Technology} \\
		\textit{Technical University of Munich}\\
		Garching bei M{\"u}nchen, Germany \\
		diermeyer@tum.de}
}

\maketitle

	\begin{acronym}
	\acro{rnd}[R\&D]{research and development}
	\acro{cots}[COTS]{commercial off-the-shelf}
	\acro{ad}[AD]{Automated Driving}
	\acro{av}[AV]{Autonomous Vehicle}
	\acro{odd}[ODD]{Operational Design Domain}
	\acro{ar}[AR]{Augmented Reality}
	\acro{qa}[QA]{Quality Assurance}
	\acro{ros}[ROS]{Robot Operating System}
	\acro{cnn}[CNN]{Convolutional Neural Network}
	\acro{rd}[RD]{Remote Driving}
	\acro{ra}[RA]{Remote Assistance}
	\acro{rm}[RM]{Remote Monitoring}
	\acro{tod}[ToD]{Teleoperated Driving}
	\acro{ipp}[IPP]{Interactive Path Planning}
	\acro{cpp}[CPP]{Collaborative Path Planning}
	\acro{wg}[WG]{Waypoint Guidance}
	\acro{sw}[SW]{Software}
	\acro{mrm}[MRM]{Minimal Risk Maneuver}
	\acro{dcpp}[DCPP]{Dynamic Collaborative Path Planning}
	\acro{hmi}[HMI]{Human-Machine Interaction}
\end{acronym}

	\begin{abstract}
Given its increasing popularity in recent years, teleoperation technology is now recognized as a robust fallback solution for \ac{ad}. \ac{ra} represents an event-driven class of teleoperation with a distinct division of tasks between the \ac{av} and the remote human operator. This paper presents a novel approach for \ac{ra} of \acp{av} in urban environments.
The concept draws inspiration from the potential synergy between highly-automated systems and human operators to collaboratively solve complex driving situations. Utilizing a hybrid algorithm that makes use of the \ac{odd} modification idea, it considers actions that go beyond the nominal operational space. 
Combined with the advanced cognitive reasoning of the human remote operator the concept offers features that hold potential to significantly improve both \ac{ra} and \ac{ad} user experiences.
\end{abstract}


	\section{Introduction}
\ac{ad} has emerged as the key technology of tomorrow’s mobility.
The transition from human-operated to self-driving vehicles promises numerous benefits due to its potential to significantly enhance transportation efficiency, safety and accessibility, thus fostering a more sustainable and efficient transportation ecosystem. However, despite the rapid advancements in different aspects of technology, \acp{av} still encounter a diverse and dynamic range of real-world driving scenarios that challenge their autonomous decision-making capabilities or even go beyond their \ac{odd}. Given its increasing popularity in recent years, teleoperation technology is now recognized as a robust fallback solution for the \ac{ad}, providing a flexible and efficient approach to tackle various challenges associated with the \ac{ad} operation life cycle \cite{Majstorovic2022-px, Amador_undated-yn}.
Finally, teleoperation holds potential beyond an \ac{ad} fallback use-case, playing a significant role in the development, deployment and enhancement of \ac{ad} technology.

\subsection{Teleoperation of Automated Vehicles}
The main objective of the teleoperation lies in assisting \acp{av} to deliver a seamless driving experience by enabling human operators to remotely assume control of the vehicle, thereby ensuring an immediate response for safe operation.
Over the years, various remote interaction concepts detailing how a human operator might remotely augment or replace an \ac{ad} function have been presented in academic literature or showcased by industry players \cite{Georg2019-dt}.
Majstorovi\'c et al. \cite{Majstorovic2022-px} offered a comprehensive overview of these interaction concepts, giving their details and discussing their viability as fallback measures in relation to the technical functionalities of the \ac{ad} pipeline and their industry adoption.
Considering the degree of human intervention during teleoperation, these concepts were grouped into two distinct categories.
The first, \ac{rd} encompasses concepts where the vehicle is entirely under remote control.
Such concepts are nowadays well-researched and represented across a wide range of use-cases.
Conversely, \ac{ra} is an event-driven task, invoked only under specific conditions and typically executed within a designated time frame.
\subsection{Remote Assistance of Automated Vehicles}
As the name suggests, \ac{ra} includes a distinct division of tasks between the system and human. Here the operator's role is primarily to offer high-level support, such as in decision-making, while the system remains in charge for the driving task.
In contrast to \ac{rd}, \ac{ra} has received considerably less attention from the community. Early research attempts were motivated by the ambition of mitigating the network latency between the operator and vehicle. In these setups, remote assistance - typically provided as a path with waypoints - is processed and executed by the vehicle. This dynamic often resulted in a so-called \textit{stop-go} behavior rendering \ac{ra} less suitable for certain teleoperation applications like mining, agriculture, etc. Conversely, such task separation in case of highly-automated driving promises a great potential.

\subsection{Contribution and Structure}

This work presents a novel approach to remote assistance of \acp{av}. Combining information from various perception modalities and underlying HD-map, this approach evaluates possible \ac{odd} modifications and formulates path suggestions for remote operators to choose from. Each suggestion is given in the form of a path accompanied by information regarding the specific \ac{odd} modification(s). This allows operators to support the \ac{av} with effective decisions, overseeing only the \ac{odd} changes, while the \ac{av} remains in charge of everything else, both driving task and overall safety.

The structure of the paper is outlined as follows: the following Section \ref{sec2} introduces related work in \ac{ra} providing more information on theoretical background. Section \ref{sec3} outlines the problem formulation. The proposed remote collaboration concept is introduced in Section \ref{sec4}, leading into Section \ref{sec5} where the proof of concept and simulation results are showcased. Finally, Section \ref{sec6} discusses future work and gives the closing remarks.

	\clearpage
\section{Related Work}
\label{sec2}

Remote assistance aims at completely decoupling the human operator from the driving task. Here, the operator assists the vehicle at the planning (tactical) level of the \ac{av} pipeline, focusing on decision-making and/or path planning \cite{Flemisch2016-fn, Majstorovic2022-px}.
This positions \ac{ra} as the preferred class of remote interaction for all stakeholders when it comes to the \ac{ad} operation in urban areas.
Early attempts to develop effective \ac{ra} interaction concepts were largely motivated by the challenges associated with one of teleoperation's most critical components - the network.
Fluctuating performance quality, packet losses, latency, or even total loss of communication between the operator and vehicle remained great challenges, even today. 
Adopting a shared autonomy control approach, Gnatzig et al. \cite{Gnatzig2012-bu, Gnatzig2015-wp} tried to minimize these network performance issues, proposing a trajectory-based concept for teleoperated driving.
While this did enhance resilience against network instabilities, it also placed operators into high-pressure situations, frequently leading to the \textit{stop-go} driving behavior.

\subsection{\ac{wg}}
Using \ac{wg}, the human operator provides guidance to the \ac{av} by specifying decisions in form of a path. Once provided, the vehicle uses these paths as reference points for trajectory generation and tracking. A basic \ac{wg} implementation was showcased in the project 5GCroCo \cite{Control2021-rd}. Meanwhile, companies Zoox \cite{Zoox2020-ae}, Cruise \cite{Cruise2021-en} and Motional \cite{Motional2022-ti} have demonstrated their respective approaches on public roads around the world.

\subsection{Collaborative Path Planning (CPP)}
Opposed to the idea where a human operator would create a path for the \ac{av} to follow,
Hosseini et al. \cite{Hosseini2014-ze} used past insights from Fong \cite{Fong_undated-ze} and came up with the idea that possible paths could be automatically generated and given to the operator to choose from. This approach is called Collaborative Path Planning (CPP) and uses a LiDAR-based occupancy grid map as a basis for path generation and clustering. The path creation process consists of two stages - path planning (RRT) and path optimization. Finally, a k-mean algorithm clustered the path candidates, presenting a selection to the remote operator to choose from. This greatly enhanced the overall teleoperation experience. However, the approach had a couple of drawbacks: it exclusively relied on a LiDAR-based occupancy grid for path planning without any goal position (target point) information, rendering it impractical for \ac{ad} applications.
Additionally, it was computationally extensive. The latter was improved by Schitz et al. \cite{Schitz2021-gw, Schitz2022-kf}. Here, they reused the idea given in \cite{Hosseini2014-ze} and proposed a similar 2-step path generation procedure based on modified RRT and CHOMP algorithms. Clustering was performed utilizing the DBSCAN algorithm, leading to a substantial improvement in computational efficiency. However, this concept remained incompatible for \ac{av} integration. Additionally, it is worth noting that Nissan published a patent \cite{Pedersen_et_undated-su} presenting ideas on what a \ac{ra} might look like. Finally, NASA's Intelligent Robotics Group demonstrated similar concepts through various project activities \cite{Intelligent_Robotics_Group_undated-ou}. However, these concepts lacked the level of detail that would allow for direct comparison with other concepts mentioned so far.

	\newpage
\section{Problem Formulation}
\label{sec3}

\begin{figure}[!b]
	\vspace*{-6mm}
	\centering
	\subfloat[Nominal \ac{odd}. Green areas represent the drivable map segments. Red areas are areas beyond the nominal \ac{odd} space.]{%
				\includegraphics[clip,width=1\columnwidth]{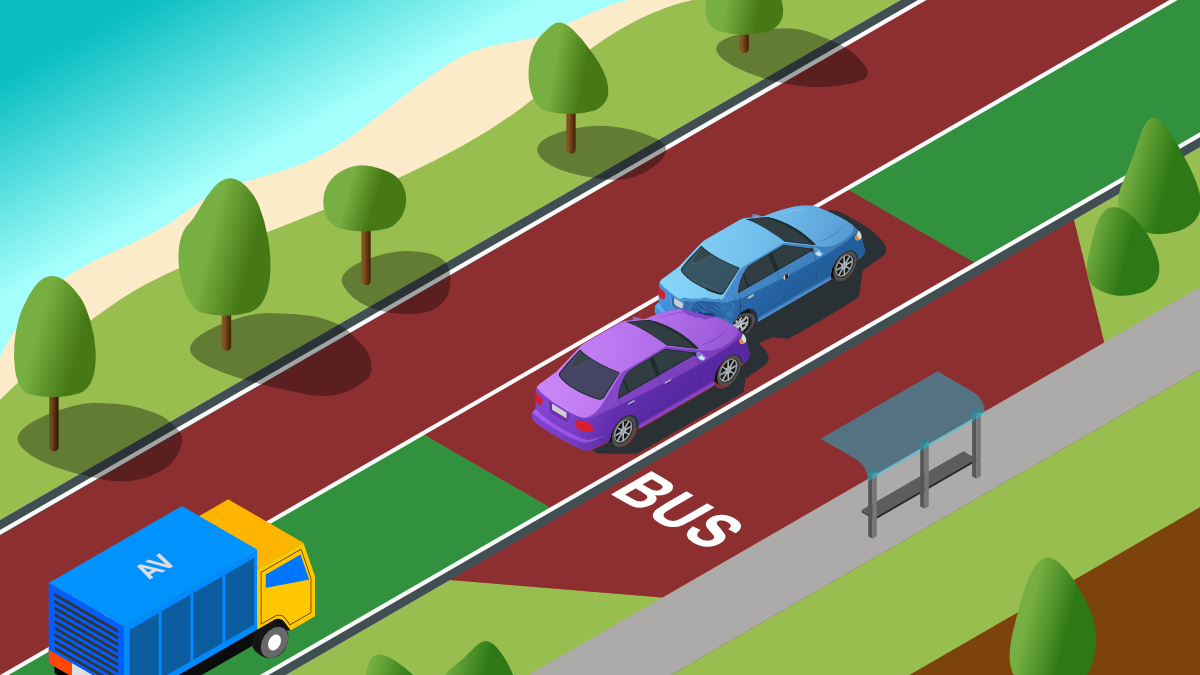}%
		\label{fig:odd_1}
	}
	\\
	\vspace*{-2mm}
	\subfloat[Modified \ac{odd}. Yellow areas represent the modified map segments, now available for planning.]{%
				\includegraphics[clip,width=1\columnwidth]{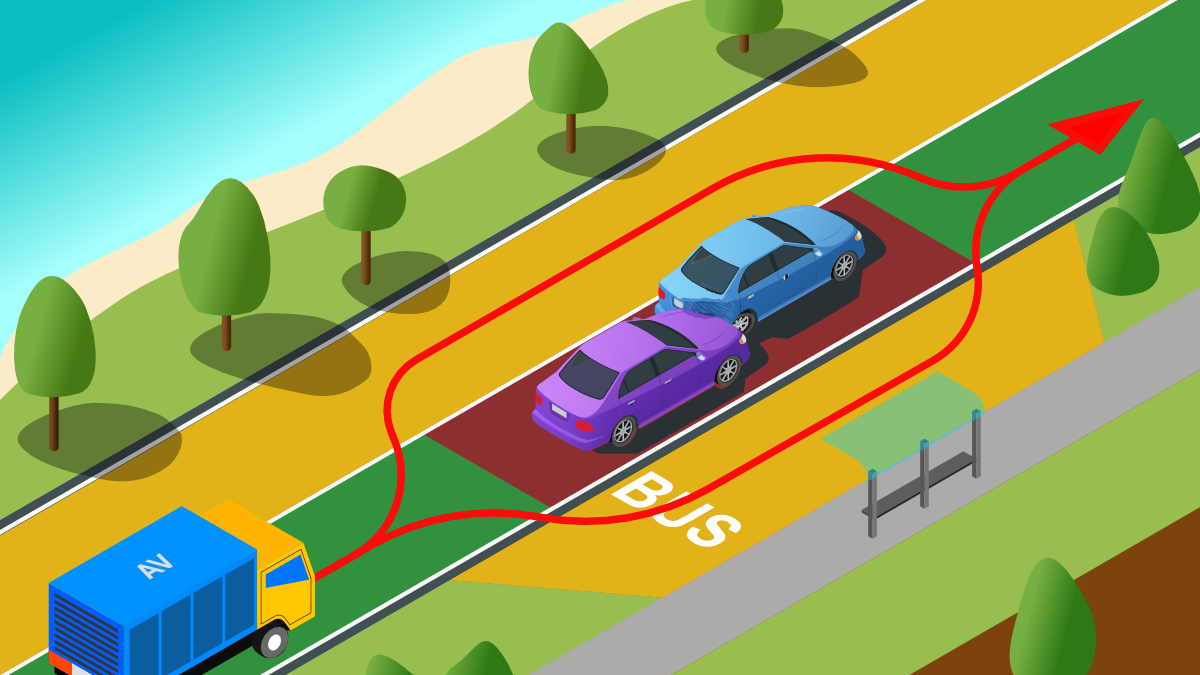}%
		\label{fig:odd_2}
	}
	\caption{Visualization of the nominal and modified (expanded) \ac{odd} forming different drivable areas for the \ac{av}}
	\label{fig:odd}
\end{figure}

The \ac{odd} of the \ac{av} defines the specific conditions under which the vehicle can operate safely. This includes a wide range of elements, from road specifics and environmental factors to traffic regulations, geographical areas, and other relevant parameters that the vehicle must respect to ensure safe and efficient operation \cite{Thorn2018-iy}. However, autonomous driving in urban settings poses a great challenge. When a vehicle approaches the boundaries of its \ac{odd} (see Fig. \ref{fig:odd_1}), it will disengage to remain compliant and seek assistance \cite{Dmv2020-di}. At this point, a human operator steps in to remotely support the vehicle at the tactical level (decision-making and/or path planning), taking full accountability for the rationale and legality of any modifications\footnote{This means that some \ac{odd} parameters will be temporary modified; for instance, a specific traffic rule will be ignored or a drivable area will be expanded beyond conventional road lanes.} to the \ac{odd} space. This interaction enables the vehicle to smoothly transition back to its nominal \ac{odd}, resuming its autonomous journey. The \ac{ra} concepts discussed earlier presented different assistance methods, often without clear insight into how decisions or paths would affect the nominal \ac{odd} space. Moreover, while the presented \ac{cpp} concepts offered notable improvement over \ac{wg} methods, they lacked seamless integration within the \ac{ad} framework. These concepts largely relied on limited data from vehicle sensors (LiDARs and cameras), producing paths that weren't fully aligned with the current \ac{av} mission. This shortcoming significantly reduces their relevance in the context of modern \ac{ra}, unveiling a substantial research gap with a great potential.

	\clearpage
\section{Approach}
\label{sec4}
\begin{figure*}[!b]
	\vspace*{-5mm}
	\centering
				\includegraphics[trim={0 0mm 0 0mm},clip,width=1\textwidth]{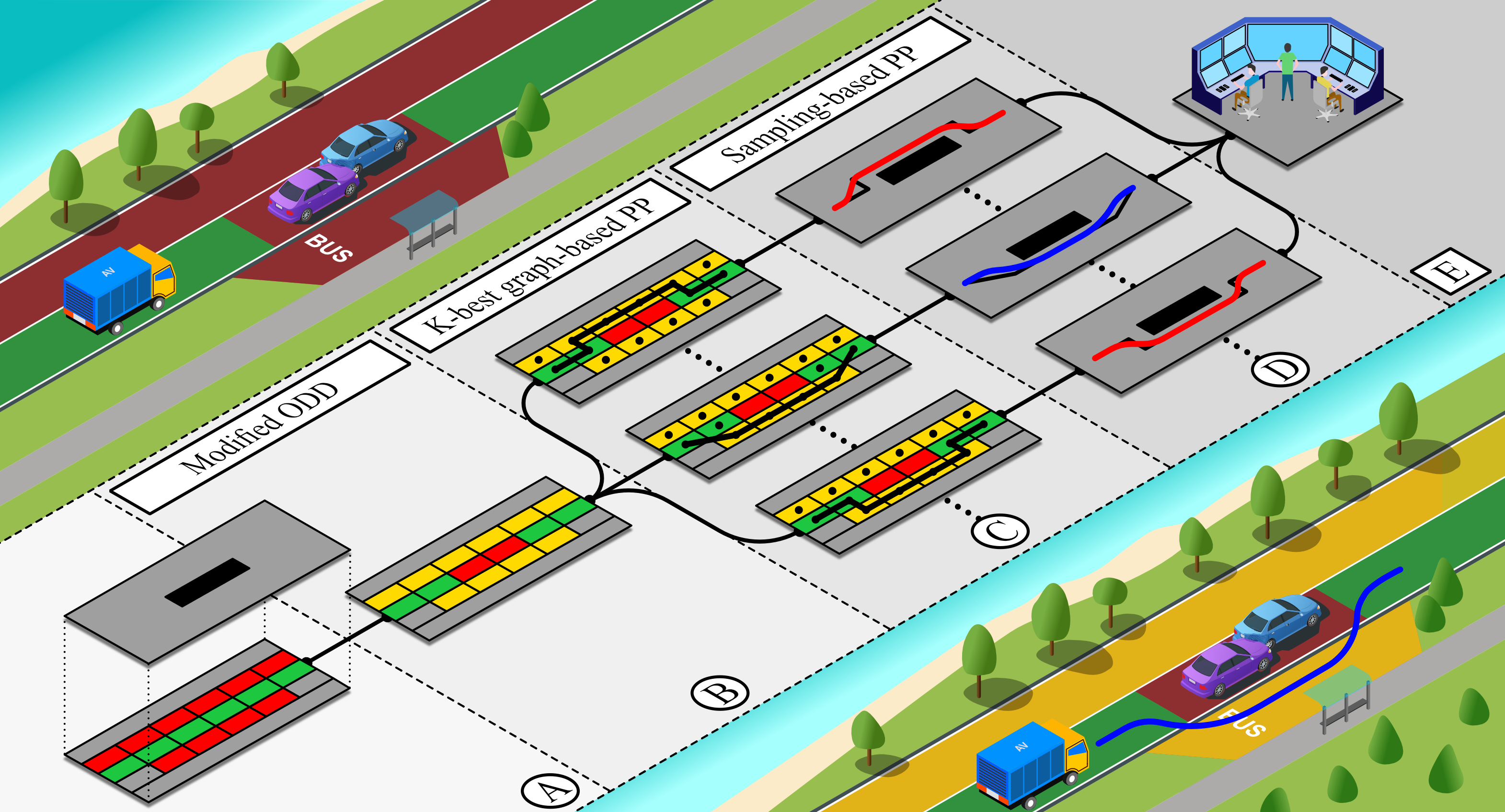}%
		\label{fig:storm_7a}
	\\
	\caption{DCPP remote interaction concept at a glance}
	\label{fig:iav3}
\end{figure*}

The problem formulation given in the previous section outlines the requirements for the novel \ac{ra} concept. In order to offer seamless and feasible integration with the \ac{av} in urban driving scenarios, the concept must:
\begin{enumerate}
	\item be developed as a functional component of the \ac{ad} pipeline - utilizing existing functional elements (perception, localization and mapping, control) and generating data that directly supports the autonomous driving task.
	\item compute and provide a remote human operator with path candidates in a \ac{cpp} manner (if applicable). Each path must contain explicit information on \ac{odd} modification(s) that took place to make the path possible.
	\item recommend a path to the operator based on specific optimization metrics such as path length, proximity to obstacles, number of ODD modifications, etc. The candidate that minimizes the cost function is proposed as the primary choice, though operators can select alternatives if they find them more appropriate.
\end{enumerate}
Finally, Table \ref{tab:stakeholders} gives details on the separation of roles and responsibilities between the two main stakeholders of the novel \ac{ra} approach.
\begin{table}[!b]
	\vspace*{-6mm}
	\caption{Task and responsibility separation between the two main stakeholders}
	\label{tab:stakeholders}
	\centering
	\begin{tblr}{
			stretch=1.25,
			colspec={|Q[c,1.5cm]|Q[l,3.0cm]|Q[l,2.75cm]|},
			rowspec={Q[m]Q[m]Q[m]}}
		\hline[1pt]
		\textbf{Stakeholder}  & \textbf{Task} & \textbf{Responsibility}  \\
		\hline[1pt]
		Autonomous Vehicle & 
		{\tabitem Path generation \\ 
		\tabitem Path tracking} & {
		\tabitem Driving Task \\
		\tabitem Safety} \\ \hline
		Human Operator &
		{\tabitem Decision-making \\ \hspace*{4mm}(behavior \& ODD)\\
		\tabitem Vehicle monitoring} & {
		\tabitem Legality of \ac{odd}\\ \hspace*{3mm} violation}\\
		\hline[1pt]
	\end{tblr}
\end{table}

\subsection{\ac{dcpp}}
To meet the requirements formulated in the previous paragraph, the novel concept of \ac{ra} called "ODD modification" is designed as follows:
When the vehicle cannot continue its mission without breaching its nominal ODD space, it disengages its \ac{ad} operation. At this point, the vehicle reaches out to the control center for remote aid. Concurrently, it updates (temporarily) the HD-map to include all obstructive entities in its vicinity and expands the drivable area by activating a predefined expanded \ac{odd} space. A graph-based pathfinding algorithm with a custom cost function sets out to find path candidates leading to the target.
If candidates exist, up to $k$ best paths are chosen as waypoints for sampling-based path planning suitable for non-holonomic dynamical systems using the present occupancy grid. These generated paths are then sent to the human operator, together with information regarding which \ac{odd} parameters were modified for each path. Additionally, the path with the minimum cost score is highlighted as the primary option. Using this information and other sensor feedback, human operator selects the path and corresponding \ac{odd} modification(s), which (if valid) the \ac{av} accepts, see Fig. \ref{fig:odd_2}. It is important to note that the vehicle retains full authority over driving and safety, possessing the capability to trigger a \ac{mrm} at any point and re-ask for remote assistance. The human operator assists the vehicle at the decision-making level and assumes accountability for the violation of the nominal \ac{odd}. Once the disengagement triggering event has been resolved, the vehicle transitions from remotely-assisted back to autonomous driving mode within its nominal \ac{odd}.

The name \ac{dcpp} highlights the real-time nature of the interaction concept, with "dynamic" emphasizing the ability to temporarily modify the \ac{odd} and "collaborative" highlighting human-machine synergy to reach a common objective.

\newpage
\subsection{DCPP Algorithm}
The following introduces algorithmic foundations of the \ac{dcpp} (see Algorithm \ref{alg1} below) as well as insights on different design decisions. The description is aligned with the topology previously presented in the Fig. \ref{fig:iav3}.

\subsubsection{High-definition Map}
The high-definition map (HD-map) plays a crucial role for safely guiding \ac{ad} systems in real-world driving environments. These maps are designed to provide a detailed representation of the road network and its surroundings encoding all relevant information to support the chosen \ac{odd}. An open-source Lanelet2 framework \cite{Poggenhans2018-vm} was selected as the HD-map framework of choice. To support the process of path planning in the subsequent step, the baseline Lanelet2 map was aligned and updated with the occupancy grid data to include objects that (currently) block lane(s) and potential alternative routes (see \krug{A}). These modifications are only temporary, and will be reverted once remote assistance ends and the vehicle returns to autonomous mode.

\algrenewcommand\algorithmicrequire{\textbf{Input:}}
\algrenewcommand\algorithmicensure{\textbf{Output:}}
\setlength{\floatsep}{10pt}
\setlength{\textfloatsep}{10pt}
\setlength{\intextsep}{10pt}

\begin{algorithm}[!b]
	\caption{DCPP algorithm}
	\label{alg1}
	\begin{algorithmic}[1]
		\Require Vehicle states, $\mathcal{S}$; HD-Map, $\mathcal{M}$; Occupancy Grid~$\mathcal{G}$; Extended ODD, $\mathcal{O_+}$
		\Ensure Path candidates, $\mathcal{C}$ coupled with the corresponding ODD modification information, $\mathcal{O_+}$ 
		\Procedure{DCPP}{$\mathcal{S}, \mathcal{M}, \mathcal{G}, \mathcal{O_+}$}
		\State $s_{RAN} \gets \text{remoteAssistanceNeeded}(\mathcal{S})$
		\Repeat
		\State $C \gets \left\langle \text{FindPathCandidates}(\mathcal{M}, \mathcal{G}, \mathcal{O_+})\right\rangle $
		\State $\text{sendCandidates}(C)$ \Comment{\textit{To operator}}
		\State $R \gets \text{receiveAssistance}()$ \Comment{\textit{From operator}}
		\If{$\text{isValid}(R)$}
		\State $\text{executeMotionControl}(R)$ 
		\Else
		\State \textbf{raise error:} "Assistance not valid"
		\State \textbf{continue} \Comment{\textit{Retry}}
		\EndIf
		\Until{$\neg s_{RAN}$}
		\EndProcedure
		\Statex \dotfill
		\Function{FindPathCandidates}{$\mathcal{M}, \mathcal{G}, \mathcal{O_+}$}
		\State $\mathcal{M} \gets \text{updateMap}(\mathcal{G})$
		\State $\mathcal{D} \gets \text{calculateDrivableArea}(\mathcal{M},\mathcal{O_+})$
		\State $R_m \gets \left\langle \text{calculateRoutes}(\mathcal{D},m) \right\rangle$ \Comment{\textit{m-best routes}}
		\If{\( |R_m| \neq 0 \)}
		\ForAll{$i \in R_m$}
		\State $Q(i) \gets \left\langle \text{calculatePath}(i), \text{paramODD}(i) \right\rangle$
		\LineCommentCont{\textit{paramODD}($i$) \textit{stores} $\mathcal{O_+}$ \textit{ODD parameters along the} $i$\textit{-th path candidate as an information to be shared with remote operators}}
		\EndFor
		\State \Return $Q$
		\Else
		\State \textbf{raise error:} "Zero candidates found"
		\EndIf
		\EndFunction			
	\end{algorithmic}
\end{algorithm}

\subsubsection{Modified \ac{odd}, $\mathcal{O}_+$}
The world's leading open-source software project for autonomous driving, Autoware \cite{The_Autoware_Foundation_undated-iy}, was taken as the primary reference for the process of the \ac{odd} definition. Here, the nominal \ac{odd}, $\mathcal{O}_n$ was defined w.r.t. vehicle type and its use case. It includes \ac{odd} parameters divided into six classes \cite{Thorn2018-iy}, where only regular roads were considered to be drivable. However, given that many disengagement scenarios often involve obstructed lanes \cite{Dmv2020-di}, and considering that human drivers typically solve such situations by driving carefully around obstacles utilizing areas not categorized as standard roads (like bus driveways, parking spaces, sidewalks, or gravel road) or occasionally (rightfully) breaching traffic rules (such as crossing a solid line), the expanded \ac{odd}, $\mathcal{O}_{+}$ includes these parameters now modified to be considered as drivable elements (see \krug{B}). Additionally, each parameter can have assigned non-negative preference value based on its priority level. For instance, driving on a bus driveway is preferred over a sidewalk. However, if only the sidewalk is available, it might be preferred as the best route. It is important to note that the \ac{av} will consider driving over such areas only under direct instruction and supervision from a human operator. The expanded \ac{odd} $\mathcal{O}_{+}$ includes modified \ac{odd} parameters from the following categories:
\begin{itemize}
	\item Physical infrastructure (including roadway types, surfaces, edges and geometry)
	\item Operational constraints (like traffic rules)
	\item Zones (comprising geo-fenced and traffic management zones)
\end{itemize}
In contrast, the \ac{odd} categories concerning objects, environmental conditions and connectivity remained unchanged.

\subsubsection{Phase I: A graph-based Path Planning}
The modified HD-Map, along with the extended \ac{odd} parameters, $\mathcal{O}_+$ forms the basis for the first phase of path generation (see \krug{C}). The graph search problem includes not only the distance but preference coefficients, too. This can be formulated as a multi-objective optimization problem with the goal of, for instance, finding a set of Pareto optimal solutions. However, to simplify the process, here the distance and preference coefficients were combined into a single cost function, $c(i, j)$ as a weighted combination of the distance, $d(i,j)$, and inverse of the specific lanelet segment preference, $p(i)$:
\begin{equation}
	c(i, j) = w_1 d(i, j) + w_2\frac{1}{p(i)}
\end{equation}

\begin{equation}
	w_1, w_2 \geq 0
\end{equation}
\begin{equation}
	\forall i, \quad p(i) > 0 
\end{equation}
$w_1$ and $w_2$ are weights used to adjust the balance of distance, $d(i,j)$, while preference, $p(i)$ of the $i$-th lanelet is given as

\begin{equation}
	p(i) = \sum_{k=1}^{n} h_k(i) \cdot p_k(i)
\end{equation}
where
\begin{equation}
	h_k(i) = \begin{cases} 
		1 & \text{if parameter $p_k$ is present in lanelet $i$} \\
		0 & \text{otherwise}
	\end{cases} \\
\end{equation}
This simplified approach aims to find the route with the best balance between these two objectives. However, having only one path candidate might not be sufficient. Some route solutions, although seemingly valid from the perspective of route planning might be rejected by the human operator as too risky or simply not aligned with human driving intuition. To increase the probability of finding a valid route that might actually be accepted by the operator, this computation phase is extended to find $k$ "best" routes.
By doing so, the combination with the underlying modified Dijkstra's algorithm \cite{LaValle2006-wf} finds $k$ path candidates for the given search problem (if exist). The target number, $k$ is usually set to 3 to prevent having too many options to choose from as they can cause additional stress or overwhelm for the remote operator \cite{Hosseini2014-ze}.

\subsubsection{Phase II: A sampling-based Path Planning}
Finding potential routes the using HD-map in the previous step is insufficient to serve as a reference path that can be tracked by the non-holonomic vehicle. Therefore, these paths are considered to be high-level planning references for the sampling-based path planning (see \krug{D}). In this second phase, the previously generated solution candidates serve as way-points for the path planning carried out by an RRT* algorithm \cite{Karaman2011-om} on the updated occupancy grid. Since the expected urban driving scenarios can often include maneuvers where a vehicle needs to change its orientation while minimizing lateral movement, the standard RRT* algorithm is augmented with the Reeds-Shepp curves \cite{Rathinam2019-qy}. This ensures support for precise maneuvering that considers the vehicle's kinematics, specifically its non-holonomic constraints. The produced paths are then sent to the remote operator, accompanied by corresponding information regarding \ac{odd} modification(s).

\subsubsection{Human-Machine Interaction}
Once the paths are generated and sent from the vehicle to the remote operator, they are displayed within the rendered 3D environment (see \kvadrat{E}). The operator assesses the situation and chooses the most appropriate path and \ac{odd} modification(s) using different interaction modalities (mouse \& keyboard, tablet with a touch screen, etc.). Ideally, the suggested path will align with the operator's preference, resulting in a swift interaction time and efficient remote assistance. Once a decision is confirmed, it is communicated back to the vehicle. Here, its operational modules generate trajectories, typically in an MPC-fashion, and further refine the path considering dynamical constrains. This ensures that the vehicle is controlled appropriately.

	\section{Proof of Concept}
\label{sec5}
The proposed \ac{odd} modification interaction concept has been developed and tested with different vehicle systems and \ac{ad} scenarios. The following briefly introduces the development phases and results obtained to date.

\subsection{Simulation Tests}
The foundational elements of the concept were developed using an open-source \ac{sw} stack for teleoperated driving based on \ac{ros}, developed and maintained by Schimpe et al. \cite{Schimpe2022-ig}.
\begin{figure}[!b]
	\vspace*{-4mm}
	\centering
	\subfloat[A \ac{dcpp} interaction scenario simulated within Carla simulator (Town03), coupled with a full stack teleoperation \ac{sw} \cite{Schimpe2022-ig}. Two path candidates that avoid the obstacle go beyond the nominal \ac{odd}. A path that minimizes the cost function (driving over parking area) is preferred and visualized with a green color. Additionally, a GUI (bottom right) provides information on corresponding ODD modifications for the selected path.]{%
		\includegraphics[trim={0mm 65mm 0mm 4mm},clip,width=1\columnwidth]{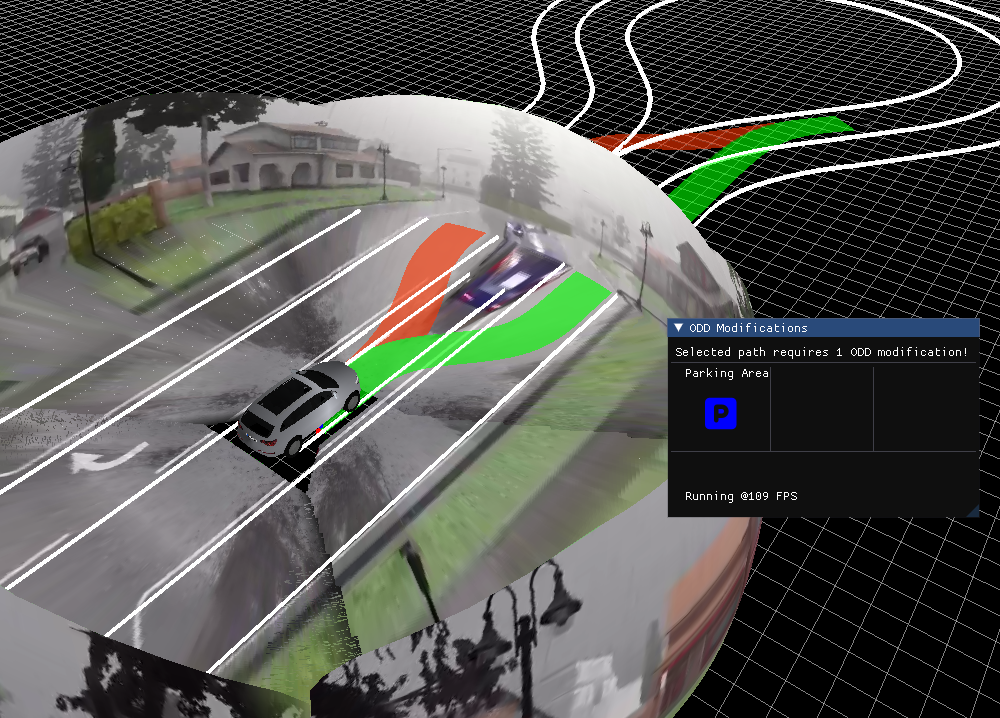}%
		\label{fig:iav4_a}
	}
	\\
	\subfloat[An aerial view of the same \ac{dcpp} interaction scenario.]{%
		\includegraphics[trim={0mm 76mm 0mm 20mm},clip,width=1\columnwidth]{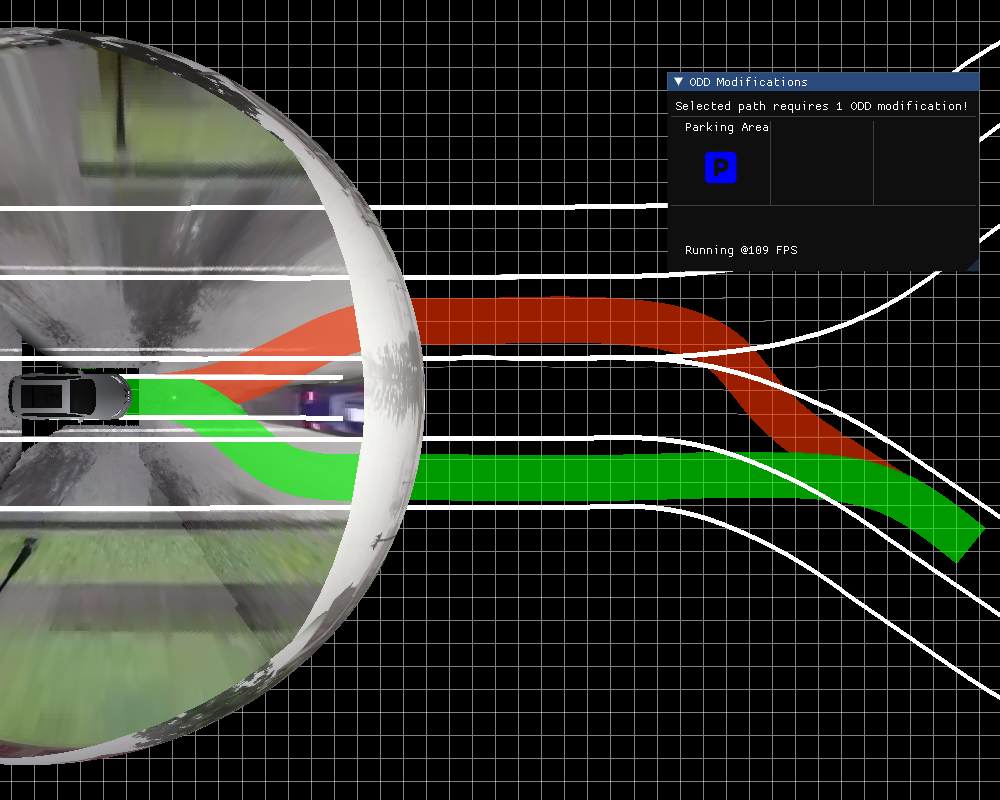}
		\label{fig:iav4_b}
	}
	\\
	\subfloat[A \ac{dcpp} interaction scenario simulated within a full-stack \ac{av} \ac{sw} \cite{The_Autoware_Foundation_undated-iy}. A single path candidate that avoids the lane obstacle (white) goes beyond the nominal \ac{odd} and suggests modifying two \ac{odd} parameters. The remote user interface visualizes affected nominal \ac{odd} parameters with icons.]{%
				\includegraphics[trim={0mm 0mm 55mm 2mm},clip,width=1\columnwidth]{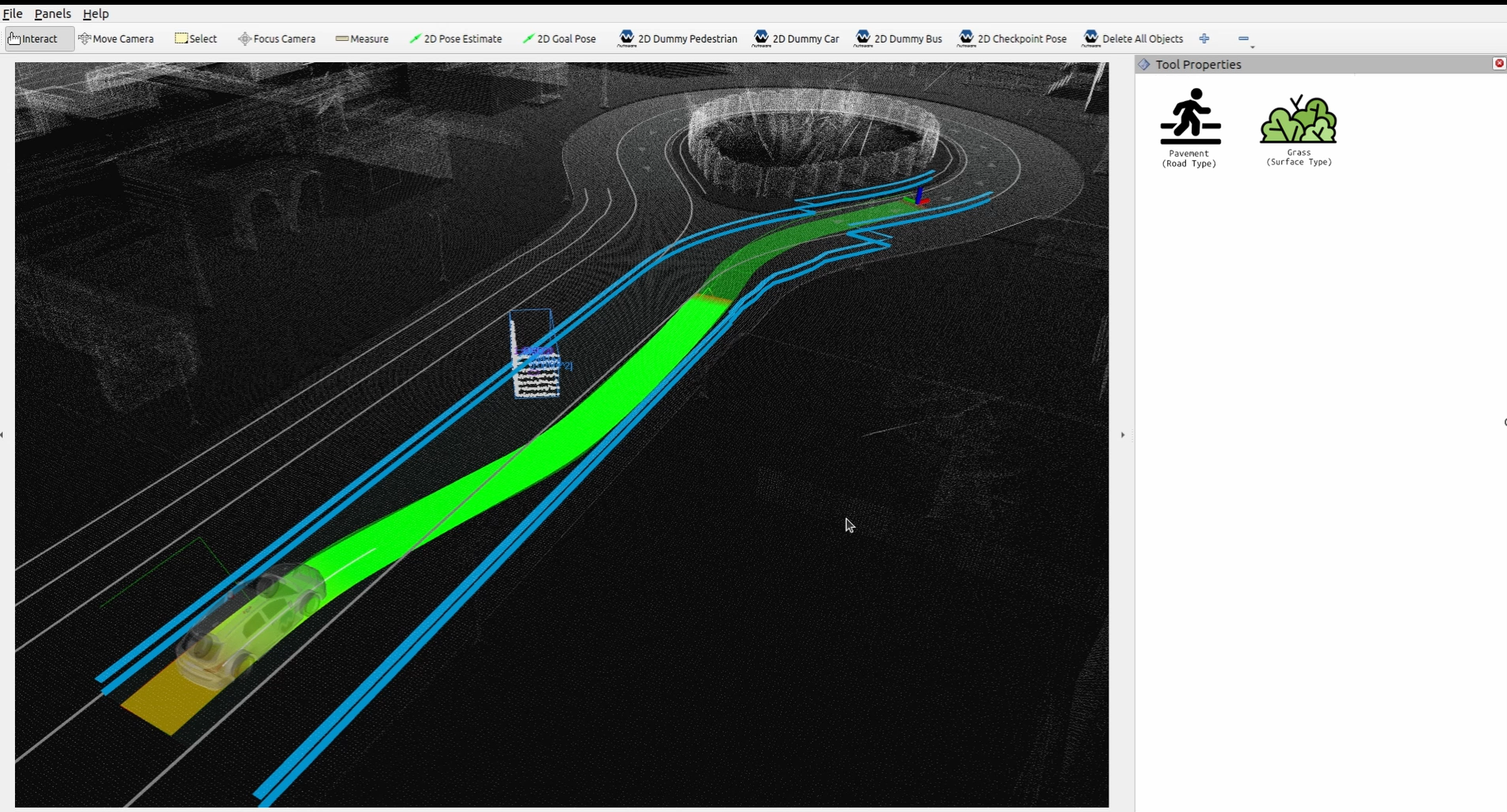}%
		\label{fig:iav4_c}
	}
	\caption{\ac{dcpp} simulation scenarios}
	\label{fig:iav4}
\end{figure}
As this \ac{sw} does not include the \ac{ad} functionality, the target position was set manually.
The rest of the concept features were implemented as described in the previous section \ref{sec4}.
The \ac{odd} parameter modifications and corresponding preference coefficients were integrated for the smaller parameter set: bus driveway, parking area, sidewalk, off-road and solid line (traffic rule violation). Preferences were prioritized in exactly the same order. Figs. \ref{fig:iav4_a} and \ref{fig:iav4_b} show the operator user interface together with the corresponding path candidates - scenario was simulated in the Carla simulator (Town03). The preferred path is highlighted in green color, and the operator can switch the selected path with keyboard keys. Corresponding \ac{odd} modification information is given as part of the GUI as well.

To simulate interactions with an \ac{av}, the proposed concept was also integrated into the open-source autoware.universe (ROS2) ecosystem \cite{The_Autoware_Foundation_undated-rx}. This has offered a platform to evaluate the capabilities of the \ac{dcpp} in interaction with an \ac{av}. Fig. \ref{fig:iav4_c} showcases the concept in action - \ac{dcpp} provides the operator with a path suggestion to drive around the static obstacle within the modified \ac{odd} space. The RViz application includes a custom panel with icons (right side of the image) displaying icons representing modified \ac{odd} parameters for a given path suggestion in the real-time.

\newpage
\subsection{Discussion and Future Work}
Unlike other \ac{ra} concepts introduced so far, \ac{odd} modification concept offers features that hold potential to significantly improve both \ac{ra} and \ac{ad} user experiences. While the current software version was tested on relatively straightforward scenarios, initial results are encouraging, suggesting that its efficacy could be maintained even in more intricate scenarios. Since the approach relies on a temporary modification (expansion) of the nominal \ac{odd}, expert workshops are necessary to further establish the parameter sets and discuss ways of the modification.
The fact that the operator does not need to construct the paths alone, indicates that \ac{ra} decisions could be streamlined (in terms of interaction time and solution quality) and might possibly significantly reduce operator stress levels. The latter, together with the assessment of the concept's effects on operator's situational awareness will be assessed in a series of human-subject studies in the near future to draw final conclusions. The development process showed the significance of user interface design with different input modalities offering a research gap worth exploring. From the algorithmic point of view, the current \ac{sw} state will be further improved and evaluated. Finally, it is expected that this contribution will also support the current activities in legislature formulation \cite{eu_undated-rt, Deutsches_Bundesministerium_fur_Digitales_und_Verkehr_BMDV_undated-hc}.

	\section{Conclusion}
\label{sec6}
This paper presented a novel approach to \ac{ra} of \acp{av} in urban environments.
The concept draws inspiration from potential synergy between highly-automated systems and human operators to collaboratively solve complex driving situations.
This way, both parties complement each other by offering strengths and receiving help as needed.
Specifically, when the vehicle reaches its operational limits and must disengage the autonomous operation, its request for remote assistance also includes suggestions on what to do next. This supports the human operator to make better decisions, and leads to efficient mission resumption and improved service.
Suggestions are the result of a hybrid algorithm that combines two path planning phases, and makes use of an \ac{odd} modification idea that considers actions that go beyond the nominal operational space.
Conversely, the human operator, with advanced cognitive reasoning and scene comprehension, instructs the vehicle on decision-making and takes responsibility for any deviation from the nominal \ac{odd}.
All these actions ensure the vehicle can be brought back into its nominal domain and eventually resume the driving mission autonomously.

	\section*{Acknowledgements}
Domagoj Majstorovi\'c as the first author, was the main developer of the presented work. Frank Diermeyer made essential contributions to the conception of the research project and revised the paper critically for important intellectual content. He gave final approval for the version to be published and agrees to all aspects of the work. As a guarantor, he accepts responsibility for the overall integrity of the paper. The authors want acknowledge the financial support for the projects UNICAR.agil (FKZ 16EMO0288) and ConnRAD (FKZ KIS8RWDS002) by the Federal Ministry of Education and Research of Germany (BMBF).

	\newpage
	\bibliography{IEEEtranBST/IEEEabrv,references}
	\bibliographystyle{IEEEtranBST/IEEEtran}
\end{document}